 \documentclass[a4paper]{article}
\usepackage{times}
\usepackage[affil-it]{authblk}
\usepackage{graphicx}
\usepackage{paralist}
\usepackage[linesnumbered,ruled,vlined]{algorithm2e} 
\usepackage{amsmath}
\usepackage{xcolor}
\usepackage{amsfonts}
\usepackage{algorithmic}
 \usepackage{sidecap}
\usepackage{times}

\title{ Artificial Prediction Markets for Online Prediction of
Continuous Variables-A Preliminary Report}

\author[1]{Fatemeh Jahedpari}
\author[1]{Marina De Vos}
\affil[1]{Department of Computer Science, University of Bath, UK}

\author[2]{Sattar Hashemi}
\affil[2]{Computer Science and Engineering Department, Shiraz University, Iran}

\author[3]{Benjamin Hirsch}
\affil[3]{EBTIC, Khalifa University, United Arab Emirates}

\author[1]{Julian Padget}

\begin{document}

\maketitle

\begin{abstract}


We propose the Artificial Continuous Prediction Market~(ACPM) as a means to predict a continuous real value, by integrating a range of data sources and aggregating the results of different machine learning (ML) algorithms. ACPM adapts the concept of the (physical) prediction market to address the prediction of real values instead of discrete events.  Each ACPM participant has a data source, a ML algorithm and a local decision-making procedure that determines what to bid on what value.  The contributions of ACPM are:
\begin{inparaenum}[(i)]
\item \emph{adaptation\/} to changes in data quality by the use of learning in:
\begin{inparaenum}[(a)]
\item the market, which weights each market participant to adjust the influence of each on the market prediction and
\item the participants, which use a Q-learning based trading strategy to incorporate the market prediction into their subsequent predictions,
\end{inparaenum}
\item \emph{resilience\/} to a changing population of low- and high-performing participants. 
\end{inparaenum}
We demonstrate the effectiveness of ACPM by application to an influenza-like illnesses data set, showing ACPM out-performs a range of well-known regression models and is resilient to variation in data source quality.\footnote{A  similar version of this work will is accepted and presented  in TRI 2015 workshop (Held at the International Joint Conference on Artificial Intelligence 2015 (IJCAI-15)). The workshop proceedings can be accessed via http://www.ufrgs.br/tri2015/}

\end{abstract}

\section{Introduction}\label{Introduction}

Physical world prediction markets aim to utilise the aggregated ``wisdom of the crowd'' to predict the outcome of a future event \cite{Ray2006}, such as who will win an election. In these markets,  participants buy and sell instruments, called securities, whose payoffs are tied to the occurrence of the specified future event. A prediction market is run by a market-maker who interacts with traders to buy and sell the securities.  
 Artificial Continuous Prediction Market (ACPM) adapts the concept for the purpose of predicting a real value in a continuous domain.  
%
%
Our motivation in developing ACPM is to use online learning in situations in which it is desirable to integrate data dynamically from a variety of sources whose data quality is (time-)variable, using a variety of analysis algorithms.


A prediction market is created for each prediction that a participant can make based on the data in their streams. All the data needed for this, including the correct prediction, is referred to as record in accordance with the ML literature. The participants, which we refer to as agents, predict the value of the record using data from their assigned source and their analysis algorithm.  Subsequently, the market maker calculates the market prediction by combining all the individual predictions.  Once the true value of the record is known, the market maker computes the reward for each agent and informs the agents about the outcome so they can 
update their analysis algorithm and their trading strategy, with the aim of improving future market predictions.  


We use a series of experiments over an influenza-Like Illness (ILI) dataset to show how ACPM can effectively be applied to the problem of syndromic surveillance.  The main objective and challenge of a syndromic surveillance system is the earliest possible detection of a disease outbreak within a population. Much research has been done to  discover potential  data sources and alternative analysis algorithms for each data source in the syndromic surveillance domain \cite{chen2010infectious}. 
An issue with syndromic surveillance data sources  is that data quality fluctuates  over time. For example, Google Flu Trends may show false alerts as a result of a sudden increase in  ILI related queries due to unusual events, such as a drug recall for a popular cold or flu remedy \cite{ginsberg2008detecting}. Therefore, integrating available data sources according to an adaptive weighting scheme over time seems necessary. In addition, given that the quality of data changes over  time, and the most suitable  algorithm for a given data source is not necessarily known \emph{a priori}, a reasonable response is to  analyse each data source with a variety of algorithms and integrate their results.  

In the experiments, we predict the level of ILI activity for a specific date in a certain region using ACPM to integrate the various data sources, analysed by different algorithms. We show that the system performs at least as well as all the market participants
and  adding learning to the agents' trading strategy improves market prediction. The results also highlight that ACPM outperforms well-known regression models  and ensembles, that are commonly used for this type of reasoning. 
The rest of the paper is organised as follows. Section~\ref{Model Description} explains the details of ACPM. Section~\ref{Evaluation} evaluates our model and analyses the results. Section~\ref{Related Work and Conclusion} covers related work and concludes.


\section{ACPM Description}\label{Model Description}
\subsection{Overview}
ACPM is an online machine learning technique which adapts the concept of a (physical) prediction market to populate it with artificial agents as market participants\footnote{The terms participating agent and agent are used interchangeably.}. We assume participants are benevolent and self-interest is not an issue, which means they are not competitive and they work together to get the best outcome of the system.
Each participating agent receives information from its designated data source and analyses its data with its given analysis algorithm. Each ACPM also includes a market maker who runs the market, deals with agent transactions and establishes the market prediction.

The market maker  instantiates a prediction market for each record 
with the purpose of predicting its true value.  Each market comprises a number of rounds,  where each agent sends its bids to the market maker. Each bid comprises:
\begin{inparaenum}[(i)]
\item a prediction value which, in our case study would be the number of cases of flu in the USA for a certain week of the year and
\item the amount the agent is betting on its prediction.
\end{inparaenum} 
Each agent, using the data for that record and its accumulated knowledge, analyses the data and predicts the true value of the record. 
Then, based on its trading strategy and its (available) capital, it determines how much to invest.  Once a round is completed, the market maker announces the market prediction based on bids received and an agent can use this information to update its bid via its trading strategy in subsequent rounds.  The market maker then seals the bid in the last round, i.e deducts capital from the agent according to its bid, rewards agents and reports the final market prediction. In this way, the period between the first round  and the last round can be used to train the agents to increase their prediction accuracy based on the integrated predictions of other participants.

Once the market is over, agents are notified of the correct answer (the true value of the record) and receive an amount of revenue as determined by a reward function.  Each agent learns from each market, based on the revenue they receive and the losses they make, in addition to finding out the correct answer.  Consequently, they can, if desired, update their strategy, analysis algorithm and beliefs for future markets.
Agents learn by updating their  analysis algorithm with the correct answer for the record and updating their trading strategy based on how much they could earn if behaving differently  (as explained in Section \ref{Agent Trading Strategy}). The market maker learns indirectly through updating the agents' capital. Their capitals determine their bidding power and hence the weight of their prediction.
The market maker integrates agent predictions using an integration
function and rewards agents based on a reward function. 
In our continuous variable prediction setting, the existing discrete
existing Market Scoring Rule (MSR) technique \cite{hanson2003combinatorial} is not suitable for our
system. In the next sections, we propose our continuous versions.

\subsection{Integration  Function}\label{Integration  Function}

At the end of each round, the market maker uses an integration function to decide the market prediction, based on the received bids. We use the following formula:

\begin{equation}\label{eq:Integration1}
\text{Market Prediction}= \frac{\sum_{b=1}^{n} Prediction_b * Invest_b}{\sum_{b=1}^{n} Invest_b}  
\end{equation}
  $$ \text{where } n \text{ is the number of bids}$$ 

\noindent
This formula assigns more weight to predictions backed by higher investments. Participants who accrue more capital, due to their success in earlier markets, have the opportunity to invest more and so get greater influence in the market. 

\subsection{Reward Function}\label{Rewards Function}

\begin{figure}[!t]
 \centering
 \includegraphics[scale=0.5]{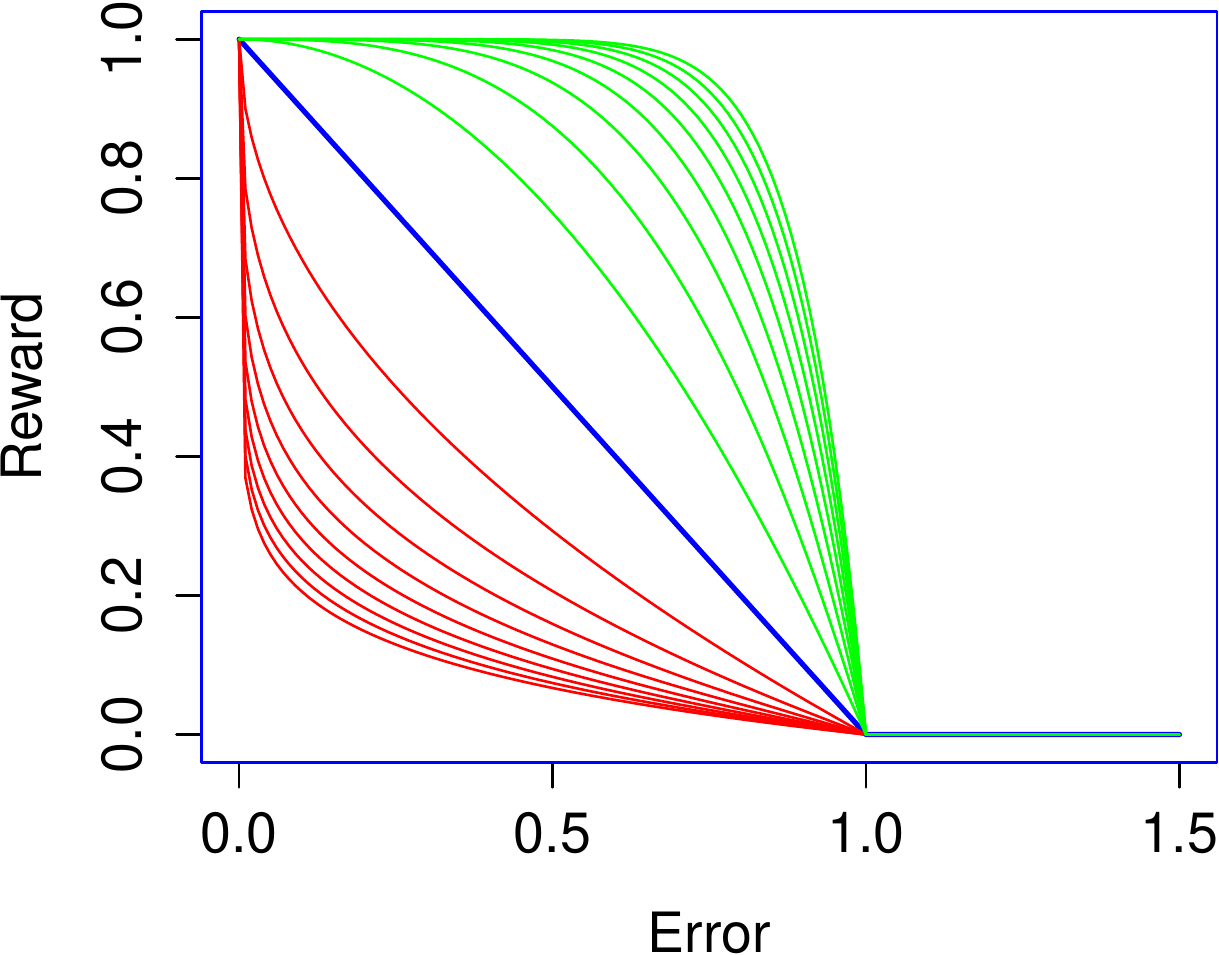}
 \caption{Reward functions. In this figure $C=1$, $\beta=1$ and each of the curve lines refers to one value of  $P \in ( \frac{1}{10},\frac{1}{9},...\frac{1}{2},1,2...10)$.}
 \label{figure:RewardFunctions}
\end{figure}

At the last round, agents are notified of the correct answer and receive revenue as determined by a reward function. These revenues are added to their capital. The reward an agent  receives is inversely proportional to the agent's prediction error, thus incentivising accurate prediction, making our reward function incentive compatible.  Equation~\ref{eq:rewardFormula} describes a family of reward functions, where different values of $P \in \mathbb{R^+}$, $\beta \in \mathbb{R^+}$ and $C \in \mathbb{R^+}$ result in the curves shown in Figure~\ref{figure:RewardFunctions}, in which $P\ge1$ generates convex functions (above diagonal) and $0<P<1$ generates concave functions (below diagonal).

\begin{equation}\label{eq:rewardFormula}
Reward = \max(\beta*\frac{ -1}{C^P}* error^P+ \beta, 0)
\end{equation}

\noindent
where $error= abs(True Value - Agent Prediction) $.
The actual revenue accrued by an agent is the product of its reward and the amount invested on its prediction.  Thus, the more an agent invests, the more revenue it receives.  Consequently, agents with higher confidence are incentivised to invest more and hence have a greater influence on the market.  In addition, the agents with low capital (indicating low past performance) cannot invest and influence the market prediction as much as high performing agents, who acquire more capital over time. 

Coefficient $C$ determines the reward cut-off, above which agents receive zero.   As can be seen in Figure \ref{figure:RewardFunctions}, the reward function is flat and equal to zero after the cut-off ($C$ = 1).
The slope of the reward function differentiates between the participating  agent rewards in proportion to the error in their predictions.  Increasing $C$, while keeping the other parameters fixed, decreases the slope of the reward function, and consequently decreases differentiation.  Conversely, increasing  $C$  increases the number of agents that receive rewards.  
  An agent's error is computed relative to the correct answer for a given market. As the range of an agent's error may change from market to market (as they learn) and from domain to domain, the cut off cannot be a fixed value, but rather be calculated for each market so that a specific percentage of agents receive positive rewards. For example, $C$ can be calculated for each market to be equal to the maximum error of all agents in that market. 
Intuitively, a certain amount of differentiation  is desirable and lower or higher values of that could harm the performance of the system.  For example, high differentiation means that a few high quality agents lead the market and  predictions of the majority of agents, including good quality ones, can be under-weighted. On the other hand, low differentiation narrows the gap between the influence of high and low quality agents, so that insufficient account is taken of the more accurate agents.
 
Coefficients $P$ and  $\beta$  shrink (or enlarge) the function horizontally and vertically respectively.  Increasing $P$ increases the degree of curvature of the reward function, and  consequently, decreases the differentiation among agents especially those with low errors.  Increasing $\beta$ has the effect of a linear increase in both agent revenue and in differentiation between participants.  With $\beta=1$, an agent loses a fraction of its money according to the error they make and only in the best case, where the error is zero, do they neither earn nor lose. This value disincentives participation, since return is less than investment and the steady depletion of capital leads to their holding very little in later markets.   The default values of $P$=1 and $\beta=2$ generate a simple linear reward function which has the property of being incentive compatible.

\subsection{Rate Per Transaction}\label{Rate Per Transaction}
The system has two other parameters: Maximum Rate Per Transaction (MaxRPT) and Minimum Rate Per Transaction (MinRPT), which specify the maximum  and minimum percentage of the capital that each participant can invest. The purpose of the MaxRPT parameter is to prevent unsuccessful agents bankrupting themselves and being eliminated from the market. It is not desirable to reduce the population, because that leads to the loss of a data feed or the loss of an analysis algorithm: while qualitatively low at some point, the combination might improve again over time.  The MaxRPT parameter can be used to tune system response to the degree of environment volatility. For example, in situations where the quality of agents' data fluctuates frequently, MaxRPT should be high so that an affluent agent loses most of its capital if its error is high for a few successive markets. On the other hand, MaxRPT should be low in situations where we expect that the quality of  good agents remains good even though they may make occasional mistakes.  If $ MaxRPT<1$, an agent's capital may get very small but is not used up entirely. Hence it can invest and recover at any time, albeit slowly!. The purpose of MinRPT is to prevent the system from being  unresponsive in cases where none of the participating agents have enough incentive to invest.

\subsection{Agent Trading Strategy}\label{Agent Trading Strategy}

As mentioned  earlier, agents can use the market prediction, received from the market maker at the end of each round, to update their bids for subsequent rounds. In this paper, we examine two strategies: a constant one and a Q-Learning based one. 

\paragraph{Constant Strategy: }\label {Constant strategy} 

Agents simply dedicate a fixed ratio of their capital to bid in each round. In this paper, this percentage is equal to MaxRPT. This na\"\i{}ve strategy ignores the advantage of updating the prediction on the basis of the market prediction of the previous round.

\paragraph{Q-Learning Trading Strategy: }\label {Q-Learning based strategy} 

In reinforcement learning, agents explore their environment and learn to choose  actions that maximise their rewards. Agents are seen as finite state machines. They receive a reward for the action they take to reach another state. 
In the Q-learning algorithm \cite{watkins1989learning}, agents have a state action value function $Q(s,a)$ which estimates the expected reward for performing an action $a$ in  state $s$. A greedy policy suggests choosing the action that gives the highest expected reward in the given state. 

In our Q-learning based trading strategy, 
agents recognise their state by 
\begin{inparaenum}[(i)]
\item measuring  the difference between their prediction and the market prediction of the previous round,
\item the current round number.
\end{inparaenum}
 Here, we have just two actions.  
 The difference between these two actions is whether the agent use the market prediction as another source of information or not to change its prediction.  While the first action (PreservePr) suggests the agent ignores the market prediction of the previous round, the second one (ChangePr) suggests the agent shifts its prediction linearly by  a percentage, called $\delta$, towards  the market prediction. 
  
In both actions, the agent uses a simple betting strategy 
 which  assumes that the correct answer is equal to the market prediction of the previous round.  Based on this assumption and its prediction value, as calculated by its analysis  algorithm, the agent estimates its error  which is absolute difference of agent prediction and market prediction. Then, the agent uses the  estimated error and the reward function setting, which was used by the market maker in the previous market, to estimate its expected reward. If the expected reward is less than one, which means that the agent  earns less than what it invests, then the  betting strategy suggests the agent invests MinRPT percentage of its capital, and otherwise MaxRPT of the capital\footnote {Two other models of betting strategy were tried, but this one both maximises system performance and agent utility. 
}. 

Agents update their state action value function  once the market is over and the correct answer is revealed. Each agent revises all  States $s$, which  it was confronted with during the market period. The agent assigns the state action values for each Action $a$ in State $s$ equal to the the amount of net revenue~-- its revenue minus the investment amount~-- it could obtain by performing Action $a$ in State $s$. The agent also calculates and stores what was the best value of  $\delta$ for state $s$.   Formula \ref{eq:calDelta} linearly calculates the percentage the agent should shift its prediction towards  the market prediction, with a limit of 100 percent.

\begin{equation}\label{eq:calDelta}
\delta= \min( abs( \frac{ \text{correct answer} - \text{agent prediction}}{ \text{market prediction} - \text{agent prediction}}),1) *100\% 
\end{equation}
  
In the first market, as the agent's knowledge is void, the agent just bids the MinRPT percentage of its capital. In all other markets, the agents have no information about the market prediction in the first round, therefore they use the constant strategy. In all other rounds, agents use a greedy strategy which means they refer to their state action value function and choose the action with  the highest  state action value. 

%
%
%
%
%
%
%
%
%
%
%
%
%
%
%
%
%
%
%
%
%
%
%
%

\section{Evaluation}\label{Evaluation}

We evaluate the performance of ACPM by applying it to syndromic surveillance in the USA. In this context, the system predicts  the disease activity level of influenza-like illnesses (ILI) in a given week in the whole of the USA using publicly available data sources.  The data  used here contains more than 100 real data streams covering the period 4th January 2004 to 27th April 2014, from a variety of sources including Google Flu Trends (GFT), Centers for Disease Control and Prevention (CDC), Google Trend. 

We have used weekly Google Flu Prediction for different areas of the United States including states, cities and regions for which GFT data is available since 2004. Google Trend statistics for different terms such as ``flu'', ``fever cough sore throat'', ``flu symptoms'' 
and CDC statistics\footnote{CDC reports ILI rates with a two-week time lag. Therefore, in order to align CDC data with the other data streams used, we take the ILI rate from two weeks earlier for each week of the experiment period.} including CDC ILI rate for different age groups, 
 USA national ILI rate, total number of patients and total number of outpatient healthcare providers in ILI network 
 are used\footnote{These data can be accessed from http://gis.cdc.gov/grasp/fluview/fluportaldashboard.html}.  
 The ACPM prediction is compared against the CDC ILI rate.

We refer to data streams as having low, medium or high quality, based on their mean absolute error (MAE) as reported by several regression models.
These categories are not absolute judgements, but relative ones confirmed through the use of several classifiers  in order to cluster the data streams according to their mean absolute error (MAE) and hence identify threshold values that fall between the clusters.

\subsection{Hypotheses}\label{hypotheses}
Using two sets of experiments, we evaluate ACPM against the following hypotheses:

\begin{compactenum}[H1:]
\item  ACPM performance is higher than  its best performing agent.\label{H1} 
\item  ACPM is resilient to different proportions of low- and high-performing participants. \label{H2}
\item  Adopting the Q-learning trading strategy, compared to the constant strategy, improves ACPM performance.\label{H3}
\item  The Q-learning trading strategy encourages low quality agents to change their prediction based on aggregated prediction of other agents.\label{H4}
\item  The Q-learning trading strategy encourages high quality agents to ignore market prediction as another source of information.\label{H5}
\item  ACPM outperforms well-known regression models

 and ensembles.\label{H6}
\item  Adopting Q-learning based trading strategy improves each participating agent's performance.\label{H7}
\end{compactenum}

\subsection{Set 1}
The first group of experiments 
explores the impact of data quality on ACPM's predictive capability. 

\paragraph{Settings}\label{Settings1}

For these experiments we look at four different market types (Table~\ref{table:MarketDensity}) with different proportions of participant data stream quality.
Market type~1 comprises only agents with medium quality data.  In order to investigate how the presence of a small number of low and high quality agents affect ACPM performance, market type~2 comprises mostly medium and a few high quality data agents and market type~3  contains  mostly medium and several low quality data agents. Market type~4 contains all three kinds (a small number of low and high quality and many medium quality data agents). Each market type has 100 agents. 

In these experiments, all agents use the Q-learning trading strategy and, randomly selected, analysis algorithm, namely  
 SGD \footnote{SGD loss function is set to Squared Loss function for the purpose of performing regression.} 
algorithm.  There is no specific reason for the use of SGD: it is just one of the several used for the initial clustering.  
 The effective values of market parameters, as discussed in Section \ref{Model Description}, can experimentally be chosen by measuring the performance of the system on historical records. Experiments gave us: 
\begin{inparaenum}[(i)]
\item number of rounds $=2$, 
\item MaxRPT$=0.9$, \item MinRPT$=0.001$, 
\item $P=7$, 
\item $\beta=4$, and
\item $C$ is chosen so that $60\%$ of agents receive positive rewards.
 \end{inparaenum}

\begin{table*}[t]
\begin{center}
\scalebox{0.7}{
\begin{tabular}{lc|c|c|c|c|c|c|}
\cline{3-8}
                                  & \multicolumn{1}{l|}{}                                               & \multicolumn{2}{c|}{Low Data Quality Agents}                                    & \multicolumn{2}{c|}{Medium Data Quality Agents}                               & \multicolumn{2}{c|}{High Data Quality Agents}                                   \\ \hline
\multicolumn{1}{|l|}{Market Type} & \begin{tabular}[c]{@{}c@{}}Average Error \\ (Variance)\end{tabular} & Quantity & \begin{tabular}[c]{@{}c@{}}Average Error \\ ( Variance)\end{tabular} & Quantity & \begin{tabular}[c]{@{}c@{}}Average Error\\ (Variance)\end{tabular} & Quantity & \begin{tabular}[c]{@{}c@{}}Average Error  \\ (Variance)\end{tabular} \\ \hline
\multicolumn{1}{|l|}{Type 1}      & \begin{tabular}[c]{@{}c@{}}0.6043 \\ (0.0007)\end{tabular}          & 0        & -                                                                    & 100      & \begin{tabular}[c]{@{}c@{}}0.6042\\ (0.0007)\end{tabular}          & 0        & -                                                                    \\ \hline
\multicolumn{1}{|l|}{Type 2}      & \begin{tabular}[c]{@{}c@{}}0.6009 \\ (0.0009)\end{tabular}          & 0        & -                                                                    & 97       & \begin{tabular}[c]{@{}c@{}}0.6033\\ (0.0008)\end{tabular}          & 3        & \begin{tabular}[c]{@{}c@{}}0.5243\\ (4.79002E-05)\end{tabular}       \\ \hline
\multicolumn{1}{|l|}{Type 3}      & \begin{tabular}[c]{@{}c@{}}0.6214 \\ (0.0030)\end{tabular}          & 12       & \begin{tabular}[c]{@{}c@{}}0.7423\\ (0.0020)\end{tabular}            & 88       & \begin{tabular}[c]{@{}c@{}}0.6048\\ (0.0008)\end{tabular}          & 0        & -                                                                    \\ \hline
\multicolumn{1}{|l|}{Type 4}      & \begin{tabular}[c]{@{}c@{}}0.6198 \\ (0.0033)\end{tabular}          & 13       & \begin{tabular}[c]{@{}c@{}}0.7406\\ (0.0019)\end{tabular}            & 84       & \begin{tabular}[c]{@{}c@{}}0.6044\\ (0.0008)\end{tabular}          & 3        & \begin{tabular}[c]{@{}c@{}}0.5243\\ (4.79002E-05)\end{tabular}       \\ \hline
\end{tabular}
}
\end{center} 
\caption{Our four market types. Data streams are divided into three categories of low, medium and high quality based on their mae as determined by several regression models. Table rows describe  market types according to data quality of their participants.} 
 \label{table:MarketDensity}
\end{table*}

\begin{figure}[!tb]
        \centering
                \includegraphics[width=0.9\columnwidth]{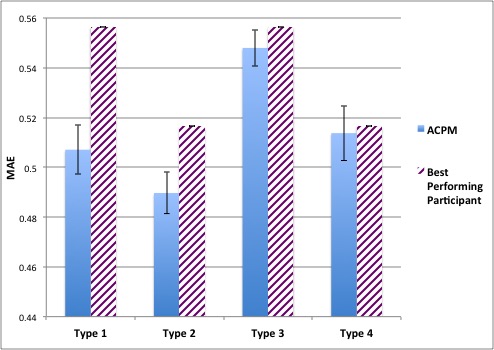}
                \caption{ Comparing ACPM performance with the best performing participant performance for each market type.}
                \label{fig:bestPerformingVsSystem}
                

\end{figure}

\begin{figure}[!tb]                
            \centering
                \includegraphics[width=0.9\columnwidth]{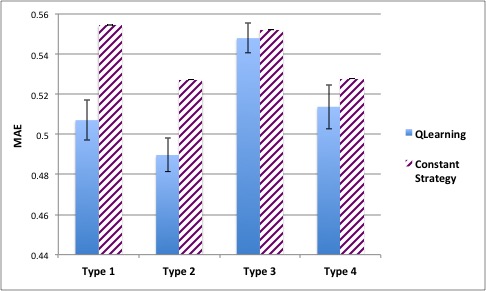}
     \caption{Comparison of ACPM's performance with Q-learning and without.}
                \label{fig:QlearningVsConstant}

\end{figure}

\begin{figure}[!tb]
                 \centering
                \includegraphics[width=0.9\columnwidth]{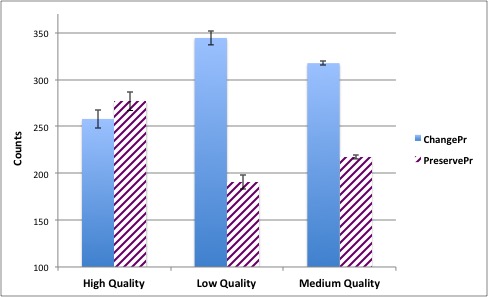}
                \caption{ Popularity of each action for agents accessing different quality of data streams. }
                \label{fig:ActionPopularity-type4}

\end{figure}

\paragraph{Experiments} 
The first experiment (Figure~\ref{fig:bestPerformingVsSystem}), compares the MAEs of the system and the best performing participant for each  market type.
Next, (Figure \ref{fig:QlearningVsConstant}), we compare, for each market type, the MAE of  ACPM  where participants use Q-learning with one where participants do not.  
In the last experiment of this set, as displayed in Figure \ref{fig:ActionPopularity-type4}, we compare 
the use of the Q-learning actions for each agent-type in a type 4 market.

\paragraph{Results}\label{Results1}
These experiments indicate that, as shown in Figure~\ref{fig:bestPerformingVsSystem}, the system's MAE is less than the best agent's MAE, without manipulating its prediction using Q-learning strategy, for every market type.   The error bars show the standard error when calculating the mean absolute error. Experiments are run once as they are deterministic. 
Figure \ref{fig:QlearningVsConstant}  shows that adopting the Q-learning reduces the MAE compared to the constant trading strategy in each market type (P-value \footnote{The null hypothesis is that the two accuracies compared are not significantly different.} $<0.05$ for all market types except Type 3).


As can be seen from Figure~\ref{fig:ActionPopularity-type4}, Action PreservePr
 which suggests the agent not change its prediction, based on the  previous round market prediction (as discussed in Section~\ref{Agent Trading Strategy}), is the most popular action in agents with high quality data and the least popular action in agents with low quality data. Conversely,  Action ChangePr which suggests the agent change its prediction by  rate $\delta$, based on the previous round market prediction, is the most popular action in agents accessing low quality data  and the least popular action in agents accessing high quality data. 


%

\subsection{Set 2}

The next group of experiments 
compares ACPM with well-known regression models
and ensembles. 

\paragraph{Settings}\label{Settings2}
 In this set of experiments, the market includes 14 participants, each agent has access to all 100 data streams  of   type 4 market, described in Table \ref{table:MarketDensity}. 
Each agent uses one of the following regression models
: SGD, IBK, LinearRegression, SMOreg, REPTree, ZeroR, DecisionStump, SimpleLinearRegression, DecisionTable, LWL, Bagging, AdditiveRegression, Stacking and Vote as its analysis algorithm.  
The market runs for two rounds and all participants use the Q-learning trading strategy. 
In these experiments, the market parameters, except $C$, are the same values as in the first set of experiments.  
Experiments indicated that, as the number of participating agents is relatively small,  $C$ is best set for each market  to maximum error so that all agents receive positive rewards. 
 Then the performance of ACPM is compared with  same models mentioned above as benchmarks.  They are run independently  without the concept of ACPM.
These models use same data as  ACPM agents do, and  similar to ACPM are run incrementally\footnote{Please note that their performance should not be compared with when they are run using batch training.}.  For each available record, they predict  the true value and then are retrained again with the correct answer and all seen records.
All models, both in ACPM and benchmarks,  are implemented using Java Weka API (3-7-10) and configured with their default parameters. 
   
\paragraph{Experiments} 

\begin{figure}[!t]
        \centering
                \includegraphics[width=0.95\columnwidth]{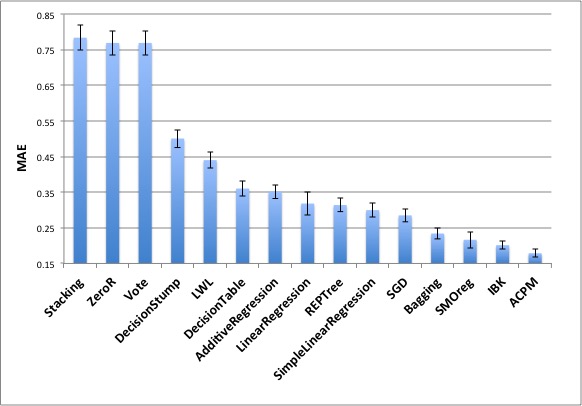}
                \caption{ Comparing ACPM performance with well known machine learning regression models 
 and ensembles.}
                \label{fig:compareWithOthers}
%
%
\end{figure}

\begin{figure}[!t]                
            \centering
                \includegraphics[width=0.95\columnwidth]{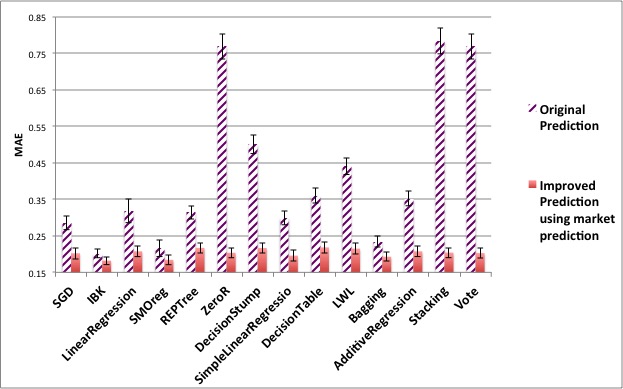}
                \caption{ How participating in ACPM and utilising Q-learning strategy  improves the performance of each classifier.}
                \label{fig:classifierImprovement}               
\end{figure}

In our first experiment (Figure~\ref{fig:compareWithOthers}) we compare ACPM's MAE with the MAE of each of the regression models 
 and ensemble methods by means of the MAE of the agents that use the method as their analysis algorithm. We then go on in Figure~\ref{fig:classifierImprovement} to compare the difference of MAE between classifier/ensemble if the agent was employing Q-learning or not. 
   

\paragraph{Results}\label{Results2}

Our experiments show (see Figure~\ref{fig:compareWithOthers})  that ACPM has a lower MAE than all regression models 
 and ensembles (P-value  is less than 0.001 for all except  IBK (P-value$=0.08$), SMOReg (P-value$=0.07$)). 
They further demonstrate that an agent using well-known regression models 
 can reduce its MAE when it uses Q-learning.


Figure~\ref{fig:classifierImprovement}  demonstrates that the performance of each classifier  is improved by participating in the market and using the Q-leaning trading Strategy (highly significant for all except  IBK and SMOReg). 

\subsection{Analysis}\label{Analyses}
A number of our hypotheses are satisfied immediately from our experiments. Given that the MAE of ACPM is always lower than the best performing agent in any market type, we can safely state that it performs better (H\ref{H1}) and that the system is resilient to different proportions of low- and high-performing participants (H\ref{H2}). Based on our first experiment, it is not surprising that ACPM performs better than regression models 
 or ensemble methods (H\ref{H6}), as demonstrated in Figure~\ref{fig:compareWithOthers}.
%
%

%
The system attains its high performance  by granting more influence to those that have high quality data sources and effective analysis algorithms.  The reward function rewards the market participants according to their prediction accuracy and the amount invested. Obviously, lower error and higher investment leads to higher revenue.  In this way, agents are incentivised to make accurate predictions and adjust their investment based on  their confidence in the prediction. After a few markets (records), the differences between agent capital becomes apparent as some of the agents gain revenue  and some of the agents loose a proportion of their capital as a result of their performance. The integration function weights each prediction by the amount of investment. In this way, higher quality agents acquire greater influence in predicting the outcome of the event, since they gain more capital over time, and consequently can invest more in their bids. 

The other reason for the performance of the system is that the agents learn to improve their prediction by considering market prediction as another source of information. Figures \ref{fig:QlearningVsConstant} and \ref{fig:classifierImprovement} show that Q-learning
does improve each agent's performance and consequently the system's performance by adding a further reduction in prediction error; hence supporting hypotheses H\ref{H3} and 
H\ref{H7}. Using the  Q-learning trading strategy, each agent learns the extent to which it should use the market prediction to update its prediction. Therefore, while high quality agents ignore market predictions, low quality agents learn to minimise the amount of noise (low accurate prediction) they send to the market maker. This is demonstrated in Figure \ref{fig:ActionPopularity-type4} and 
confirms H\ref{H4} and H\ref{H5}.
The validity of the ACPM approach through its application to several of the UCI data sets is confirmed, but  cannot be presented here due to sake of space.

\section{Related Work and Conclusion}\label{Related Work and Conclusion}

We proposed  an Artificial Continuous Prediction Market (ACPM)
for predicting a continuous variable based on the integration of diverse data sources with different varying quality. It acts as an adaptive ensemble algorithm which is capable of shifting focus in response to changes in individuals' predictions.

To our knowledge, there is relatively little research on artificial prediction markets as a machine learning technique.
Our work is different from  related works in artificial prediction markets \cite{perols2009information,barbu2012introduction,journals/jmlr/Storkey11,millin2012isoelastic,jumadinova2013prediction},  prediction with expert advice and its subfields \cite{vovk1995game,chen2010new,freund1997decision,hazan201210,littlestone1994weighted,shalev2007primal,kalai2005efficient}, opinion pools and all ensemble techniques
  as learning happens at two levels, i.e. market and agents. The market learns the  weighting of each agent   on the market prediction dynamically while participants revise their beliefs and can retrain themselves 
\begin{inparaenum}[(i)]
\item after each round of a market by comparing their prediction with market prediction to  maximise their utility in the current market.
\item  after each market in order to maximise their utility in future markets.
\end{inparaenum}
 Finally, we note that previous works are designed for discrete classification and our work is designed to predict a continuous variable.

Our next step is 
to develop an intelligent market that can self-select the appropriate parameters for the market based on the characteristics of market participants 
and their data sources.
We also plan to apply ACPM on different domains, such as, for example, stock market and cancer predictions.

\bibliographystyle{plain}
\bibliography{Main} 

\end{document}